# Benchmarking Deep Facial Expression Recognition: An Extensive Protocol with Balanced Dataset in the Wild

Gianmarco Ipinze Tutuianu*, Yang Liu*, Ari Alamäki, Janne Kauttonen
* Equal contribution

*Abstract*

Facial expression recognition (FER) is a crucial part of human-computer interaction. Existing FER methods achieve high accuracy and generalization based on different open-source deep models and training approaches. However, the performance of these methods is not always good when encountering practical settings, which are seldom explored. In this paper, we collected a new in-the-wild facial expression dataset for cross-domain validation. Twenty-three commonly used network architectures were implemented and evaluated following a uniform protocol. Moreover, various setups, in terms of input resolutions, class balance management, and pre-trained strategies, were verified to show the corresponding performance contribution. Based on extensive experiments on three large-scale FER datasets and our practical cross-validation, we ranked network architectures and summarized a set of recommendations on deploying deep FER methods in real scenarios. In addition, potential ethical rules, privacy issues, and regulations were discussed in practical FER applications such as marketing, education, and entertainment business.

## I. Introduction

Facial expression recognition (FER) allows machines to interpret human emotions using computer vision techniques through images and videos. FER is significant because it has broad use cases including intelligent marketing [1], AR/VR devices [2], mental state measurement [3], and social interactions [4].

Recent FER methods based on deep learning have achieved high recognition accuracy and good generalization ability based on large-scale annotated data and powerful computing resources [5], [6]. Nevertheless, several factors still challenge the performance of FER methods, especially for practical deployment. First, environmental diversities, such as illumination and occlusion, can negatively influence facial expressions, leading to misclassification [7], [8]. Second, low and inconsistent data quality due to camera parameters or compressions might cause heavy performance losses [9]. Third, data imbalance often occurs in public FER datasets, which lets certain emotions become over or less represented, resulting in an overfitting problem [10], [11]. To overcome the above challenges, previous works proposed solutions like collecting larger datasets, introducing multi-modal signals, applying data augmentation, and balancing techniques, respectively or collaboratively [5], [6], [8].

Nevertheless, since different methods were implemented with different backbone networks, datasets, and training strategies, it is hard to understand real sources of performance improvement [5], [9]. Alternatively, a few recent studies have delved implemental settings such as different network architectures. However, the conclusion is that a specific model family is more likely to perform better on a specific dataset, which is not helpful enough for practical deployment. There is no solid study consisting of uniform protocol and cross-domain benchmark.

To this end, we propose a comprehensive protocol for evaluating practical FER deployment. The goal of the above protocol is to answer the following research questions (RQ):

- RQ1: Whether different resolutions exhibit FER performances differences in practical settings?
- RQ2: Whether pre-trained weights increase FER performances in practical settings?
- RQ3: Whether built-in balancing strategies during the model training improve FER performance in practical settings?

In this paper, the FER performance of twenty-three popular deep models is evaluated to answer the above RQs and thus offer practical insights for FER applications, and also guiding the development of new models. The contribution of this paper is summarized as follows:

- A new balanced in-the-wild facial expression dataset, including 2,100 images with annotations of seven basic emotions, is collected for cross-domain validation.
- Twenty-three and seventeen network architectures are implemented within a uniform configuration for small-size and normal-size input, respectively.
- Extensive experiments, in terms of various input resolutions, pre-trained strategies, and class balancing managements, are conducted to address three RQs, whose findings provide recommendations of FER deployment in practical applications.

## II. BALANCED TESTING OF FACIAL EXPRESSION RECOGNITION DATASET

To facilitate training advanced deep FER models, a few datasets have been established with in-the-wild conditions, such as RAF-DB [12], FER2013 [13] and AffectNet [14]. Sample images from these datasets are shown in Fig. 1).

RAF-DB dataset consists of facial images captured in real-world scenarios, capturing a wide range of emotions expressed by individuals in naturalistic settings. It contains diverse emotions, including anger, disgust, fear, happiness, sadness, surprise, and neutrality. The images size is 100 x 100.

FER2013 dataset is widely used and consists of over 35,000 grayscale images of facial expressions belonging to seven categories: anger, disgust, fear, happiness, sadness, surprise, and neutral. The image size is 48 x 48.

AffectNet dataset contains a large collection of facial expression images captured in uncontrolled environments, covering a wide range of emotions and intensities. The dataset provides comprehensive annotations, including eight emotion labels, valence and arousal, and dominance ratings. The image size is not aligned.

Although these datasets provide large-scale annotated training data, they exist significant problems in terms of different image size and imbalanced categories. For example, the RAF-DB dataset has 887 disgust samples and 5957 happiness samples, respectively, as shown in Table 1. These in-built biases can lead to weak performance on specific classes and overfitting problems. Moreover, existing studies usually apply a pre-trained backbone to extract facial features by resizing different samples to the required resolution. There are a few reports about the impact of the resizing process on FER performance.

To this end, we collected a new dataset to fairly compare the performance of existing models with in-the-wild facial expression images in practical conditions. This gives us better insight into potential problems with our models, such as overfitting, biases, robustness, and generalizability. The novel dataset is called balanced testing of facial expression recognition, BTFER. Next, we describe the dataset establishment in detail.

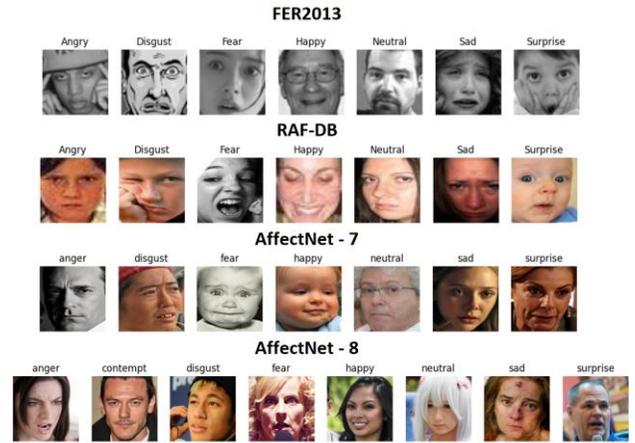

Fig. 1: Sample images form four datasets.

Stage 1: we first created a control dataset, using google images download library (https://github.com/hardikvasa/google-images-download), to collect a substantial number of images from the internet. To acquire relevant images for our face expression recognition dataset, appropriate keywords such as "white man angry, Asian men happy, black women sad, latino women surprised, Indian kid smile", were used as search queries, to have different ethnicities, ages, and genders.

Stage 2: The obtained images were then subjected to a meticulous manual cleaning procedure to ensure dataset integrity. During the manual cleaning process, various criteria were applied to filter out unwanted images. For example, non-human faces, irrelevant images, duplicates, and other elements that could introduce bias, or noise were eliminated.

Stage 3: Following the cleaning stage, the MTCNN [15] was employed to automatically locate and isolate facial regions within the collected images. By extracting and cropping the detected faces, the dataset focused exclusively on the relevant region of interest (ROI).

Stage 4: To ensure consistency and comparability across the dataset, all cropped facial images were resized to a standard dimension of 256 x 256 pixels, which includes 20 pixels over the ROI. It was necessary for context preservation, keeping facial integrity and reducing alignment issues.

Stage 5: Finally, a subset of 300 images of each category was randomly selected to build as a representative testing dataset for subsequent cross-validation experiments.

The created BTFER dataset has balanced classes and well annotated facial expression images in the wild. It was employed with our practical protocol to evaluate all the selected models in the following experiments.

| Dataset Name | Expressions | No Images per class | No. Total of Images (Dataset) | Resolution |
| --- | --- | --- | --- | --- |
| Real-world affective Database (RAF-DB) | Angry | 867 | 15 339 | 100 x 100 |
| | Disgust | 877 | | |
| | Fear | 355 | | |
| | Happy | 5957 | | |
| | Neutral | 3204 | | |
| | Sad | 2460 | | |
| | Surprise | 1619 | | |
| FER2013 (Facial Expression Recognition 2013 Dataset) | Angry | 4953 | 35 887 | 48 x 48 |
| | Disgust | 547 | | |
| | Fear | 5121 | | |
| | Happy | 8989 | | |
| | Neutral | 6198 | | |
| | Sad | 6077 | | |
| | Surprise | 4002 | | |
| AffectNet | Angry | 3353 | 25 938 | 512 x 512 |
| | Disgust | 2480 | | |
| | Fear | 3289 | | |
| | Happy | 4791 | | |
| | Neutral | 4878 | | |
| | Sad | 3164 | | |
| | Surprise | 3983 | | |
| BTFER (Test Dataset) | Angry | 300 | 2 100 | 256 x 256 |
| | Disgust | 300 | | |
| | Fear | 300 | | |
| | Happy | 300 | | |
| | Neutral | 300 | | |
| | Sad | 300 | | |
| | Surprise | 300 | | |

Table 1. Datasets used in this work with descriptions

| Model | Num Parameter | Dataset | Simple | | Conservative | |
|---|---|---|---|---|---|---|
| | | | Validation Accuracy | BTFER Accuracy | Validation Accuracy | BTFER Accuracy |
| VGG16 - IN | ~15M | RAF-DB | 0,77 | 0,52 | 0,79 | 0,52 |
| | | FER2013 | 0,68 | 0,54 | **0,68** | **0,54** |
| | | AffectNet 7 | **0,62** | **0,54** | 0,63 | 0,44 |
| VGG16 - VF | ~15M | RAF-DB | 0,72 | 0,45 | 0,75 | 0,46 |
| | | FER2013 | 0,66 | 0,48 | 0,65 | 0,50 |
| | | AffectNet 7 | 0,60 | 0,46 | 0,60 | 0,45 |
| VGG16 - SC | ~15M | RAF-DB | 0,73 | 0,47 | 0,76 | 0,51 |
| | | FER2013 | 0,67 | 0,54 | 0,68 | 0,52 |
| | | AffectNet 7 | 0,56 | 0,41 | 0,57 | 0,37 |
| VGG19 | ~20M | RAF-DB | **0,78** | **0,54** | 0,78 | 0,54 |
| | | FER2013 | **0,68** | **0,55** | 0,67 | 0,52 |
| | | AffectNet 7 | 0,63 | 0,44 | **0,64** | **0,44** |
| MobileNet | ~6M | RAF-DB | 0,72 | 0,42 | 0,72 | 0,42 |
| | | FER2013 | 0,67 | 0,41 | 0,63 | 0,41 |
| | | AffectNet 7 | 0,60 | 0,36 | 0,62 | 0,36 |
| MobileNetV2 | ~5M | RAF-DB | 0,71 | 0,41 | 0,65 | 0,37 |
| | | FER2013 | 0,61 | 0,35 | 0,62 | 0,41 |
| | | AffectNet 7 | 0,57 | 0,46 | 0,60 | 0,34 |
| EfficientNetV2_B0 | ~6M | RAF-DB | 0,61 | 0,35 | 0,68 | 0,43 |
| | | FER2013 | 0,59 | 0,47 | 0,61 | 0,43 |
| | | AffectNet 7 | 0,53 | 0,38 | 0,50 | 0,38 |
| NasNet | ~4,5M | RAF-DB | 0,68 | 0,39 | 0,67 | 0,39 |
| | | FER2013 | 0,60 | 0,30 | 0,63 | 0,29 |
| | | AffectNet 7 | 0,43 | 0,33 | 0,50 | 0,38 |
| ResNet50 | ~24M | RAF-DB | 0,68 | 0,43 | 0,72 | 0,42 |
| | | FER2013 | 0,64 | 0,39 | 0,64 | 0,45 |
| | | AffectNet 7 | 0,58 | 0,38 | 0,55 | 0,38 |
| DenseNet121 | ~7M | RAF-DB | 0,69 | 0,42 | 0,74 | 0,45 |
| | | FER2013 | 0,64 | 0,47 | 0,66 | 0,45 |
| | | AffectNet 7 | 0,57 | 0,48 | 0,36 | 0,29 |
| AlexNet * | ~21,5M | RAF-DB | 0,53 | 0,32 | 0,63 | 0,38 |
| | | FER2013 | 0,47 | 0,43 | 0,59 | 0,50 |
| | | AffectNet 7 | 0,50 | 0,43 | 0,50 | 0,38 |
| Ensemble | ~2M | RAF-DB | 0,64 | 0,40 | 0,66 | 0,40 |
| | | FER2013 | 0,61 | 0,35 | 0,63 | 0,41 |
| | | AffectNet 7 | 0,63 | 0,41 | 0,63 | 0,41 |
| Sequential 3Conv | ~1,3M | RAF-DB | 0,59 | 0,37 | 0,72 | 0,46 |
| | | FER2013 | 0,60 | 0,49 | 0,66 | 0,53 |
| | | AffectNet 7 | 0,60 | 0,39 | 0,59 | 0,41 |
| Sequential 4Conv * | ~3M | RAF-DB | 0,65 | 0,47 | 0,76 | 0,54 |
| | | FER2013 | 0,61 | 0,52 | 0,66 | 0,48 |
| | | AffectNet 7 | 0,59 | 0,38 | 0,61 | 0,40 |
| Modular | ~5M | RAF-DB | 0,63 | 0,41 | 0,72 | 0,45 |
| | | FER2013 | 0,60 | 0,49 | 0,65 | 0,53 |
| | | AffectNet 7 | 0,56 | 0,39 | 0,56 | 0,38 |
| RepVGG | ~26,5M | RAF-DB | 0,60 | 0,32 | 0,72 | 0,46 |
| | | FER2013 | 0,65 | 0,38 | 0,65 | 0,20 |
| | | AffectNet 7 | 0,58 | 0,39 | 0,58 | 0,42 |
| Sequential Simple * | ~2.5 | RAF-DB | 0,67 | 0,48 | 0,75 | 0,50 |
| | | FER2013 | 0,61 | 0,51 | 0,66 | 0,55 |
| | | AffectNet 7 | 0,53 | 0,40 | 0,46 | 0,37 |
| SMMB | ~5,5M | RAF-DB | 0,66 | 0,37 | 0,69 | 0,46 |
| | | FER2013 | 0,52 | 0,49 | 0,63 | 0,51 |
| | | AffectNet 7 | 0,52 | 0,39 | 0,46 | 0,34 |
| VGGNet | ~15M | RAF-DB | 0,61 | 0,38 | 0,77 | 0,51 |
| | | FER2013 | 0,58 | 0,31 | 0,66 | 0,54 |
| | | AffectNet 7 | 0,47 | 0,37 | 0,52 | 0,33 |
| LeNet5 * | ~0,2M | RAF-DB | 0,53 | 0,35 | 0,65 | 0,37 |
| | | FER2013 | 0,48 | 0,45 | 0,53 | 0,40 |
| | | AffectNet 7 | 0,53 | 0,33 | 0,49 | 0,32 |
| ResNet18 + Dropout + Dense | ~12M | RAF-DB | 0,74 | 0,54 | **0,81** | **0,56** |
| | | FER2013 | 0,67 | 0,49 | 0,68 | 0,53 |
| | | AffectNet 7 | 0,57 | 0,41 | 0,57 | 0,44 |
| Mediapipe Feature Extraction + VGGNet | ~5,5M | RAF-DB | 0,74 | 0,40 | 0,78 | 0,46 |
| | | FER2013 | 0,66 | 0,55 | 0,66 | 0,53 |
| | | AffectNet 7 | 0,65 | 0,39 | 0,65 | 0,41 |
| MNIST_NET | ~18M | RAF-DB | 0,71 | 0,53 | 0,79 | 0,56 |
| | | FER2013 | 0,55 | 0,47 | 0,69 | 0,53 |
| | | AffectNet 7 | 0,45 | 0,33 | 0,50 | 0,36 |
| ZF_NET | ~21,5M | RAF-DB | 0,61 | 0,40 | 0,67 | 0,39 |
| | | FER2013 | 0,56 | 0,51 | 0,61 | 0,52 |
| | | AffectNet 7 | 0,55 | 0,36 | 0,55 | 0,37 |
| GoogLeNet | ~6M | RAF-DB | 0,71 | 0,50 | 0,74 | 0,55 |
| | | FER2013 | 0,65 | 0,54 | 0,66 | 0,55 |
| | | AffectNet 7 | 0,53 | 0,42 | 0,56 | 0,44 |

Table 2: Performance comparison in models with 48 x 48 resolution images. Bolded values indicate the top-3 values for each of the three model-training dataset. * = Flattening pooling

| Model | Num Parameter | Dataset | Simple | | Conservative | |
|---|---|---|---|---|---|---|
| | | | Validation Accuracy | BTFER Accuracy | Validation Accuracy | BTFER Accuracy |
| VGG-16 - IN | ~15M | RAF-DB | 0,82 | 0,59 | 0,83 | 0,56 |
| | | FER2013 | 0,67 | 0,54 | 0,69 | 0,57 |
| | | AffectNet | 0,71 | 0,60 | 0,69 | 0,62 |
| VGG-16 - VF | ~15M | RAF-DB | 0,83 | 0,60 | 0,85 | 0,60 |
| | | FER2013 | **0,71** | **0,58** | 0,72 | 0,56 |
| | | AffectNet | 0,71 | 0,63 | 0,70 | 0,66 |
| VGG-16 - SC | ~15M | RAF-DB | 0,82 | 0,56 | 0,82 | 0,56 |
| | | FER2013 | 0,69 | 0,56 | 0,69 | 0,56 |
| | | AffectNet | 0,71 | 0,60 | 0,71 | 0,58 |
| VGG-19 | ~20M | RAF-DB | 0,81 | 0,59 | 0,82 | 0,57 |
| | | FER2013 | 0,68 | 0,55 | 0,69 | 0,57 |
| | | AffectNet | 0,72 | 0,63 | 0,72 | 0,56 |
| ResNet50 - IN | ~24M | RAF-DB | 0,82 | 0,56 | 0,82 | 0,59 |
| | | FER2013 | 0,68 | 0,50 | 0,69 | 0,52 |
| | | AffectNet | 0,71 | 0,56 | 0,70 | 0,55 |
| ResNet50 - VF | ~24M | RAF-DB | **0,86** | **0,62** | **0,87** | **0,60** |
| | | FER2013 | 0,70 | 0,56 | 0,71 | 0,61 |
| | | AffectNet | 0,71 | 0,56 | 0,72 | 0,61 |
| MobileNet | ~6M | RAF-DB | 0,78 | 0,52 | 0,81 | 0,54 |
| | | FER2013 | 0,67 | 0,48 | 0,70 | 0,52 |
| | | AffectNet | 0,71 | 0,56 | 0,71 | 0,54 |
| MobileNetV2 | ~5M | RAF-DB | 0,78 | 0,53 | 0,77 | 0,51 |
| | | FER2013 | 0,66 | 0,42 | 0,68 | 0,53 |
| | | AffectNet | 0,70 | 0,56 | 0,69 | 0,55 |
| Xception | ~21.5M | RAF-DB | 0,80 | 0,54 | 0,81 | 0,52 |
| | | FER2013 | 0,70 | 0,49 | 0,71 | 0,53 |
| | | AffectNet | 0,71 | 0,61 | 0,72 | 0,61 |
| SeNet50 | ~27M | RAF-DB | 0,85 | 0,59 | 0,86 | 0,59 |
| | | FER2013 | 0,69 | 0,54 | **0,71** | **0,62** |
| | | AffectNet | **0,74** | **0,60** | 0,69 | 0,61 |
| EfficientNetV2B0 | ~6M | RAF-DB | 0,71 | 0,57 | 0,71 | 0,58 |
| | | FER2013 | 0,54 | 0,41 | 0,66 | 0,42 |
| | | AffectNet | 0,72 | 0,58 | 0,69 | 0,56 |
| NasNet | ~5M | RAF-DB | 0,75 | 0,48 | 0,76 | 0,49 |
| | | FER2013 | 0,67 | 0,49 | 0,67 | 0,49 |
| | | AffectNet | 0,69 | 0,57 | 0,68 | 0,55 |
| InceptionV3 | ~26M | RAF-DB | 0,80 | 0,52 | 0,82 | 0,54 |
| | | FER2013 | 0,69 | 0,40 | 0,70 | 0,49 |
| | | AffectNet | 0,71 | 0,59 | 0,72 | 0,60 |
| DenseNet121 | ~7M | RAF-DB | 0,82 | 0,55 | 0,82 | 0,60 |
| | | FER2013 | 0,69 | 0,50 | 0,70 | 0,52 |
| | | AffectNet | 0,71 | 0,61 | **0,72** | **0,61** |
| RepVGG | ~34M | RAF-DB | 0,78 | 0,56 | 0,81 | 0,55 |
| | | FER2013 | 0,69 | 0,44 | 0,69 | 0,47 |
| | | AffectNet | 0,70 | 0,61 | 0,71 | 0,62 |
| Sequential Simple | ~22M | RAF-DB | 0,61 | 0,43 | 0,69 | 0,47 |
| | | FER2013 | 0,60 | 0,48 | 0,62 | 0,36 |
| | | AffectNet | 0,66 | 0,52 | 0,66 | 0,53 |
| VGGNet | ~12M | RAF-DB | 0,77 | 0,55 | 0,82 | 0,57 |
| | | FER2013 | 0,67 | 0,58 | 0,70 | 0,60 |
| | | AffectNet | 0,69 | 0,57 | 0,70 | 0,56 |
| ResNet18 | ~6M | RAF-DB | 0,77 | 0,53 | 0,82 | 0,56 |
| | | FER2013 | 0,67 | 0,54 | 0,70 | 0,63 |
| | | AffectNet | 0,68 | 0,56 | 0,69 | 0,56 |
| GoogLeNet | ~89M | RAF-DB | 0,75 | 0,60 | 0,79 | 0,58 |
| | | FER2013 | 0,68 | 0,63 | 0,68 | 0,62 |
| | | AffectNet | 0,69 | 0,58 | 0,69 | 0,61 |
| Modular | ~103M | RAF-DB | 0,66 | 0,48 | 0,73 | 0,50 |
| | | FER2013 | 0,58 | 0,49 | 0,64 | 0,53 |
| | | AffectNet | 0,67 | 0,56 | 0,66 | 0,54 |

Table 3: Performance comparison in models with 224 x 224 resolution images. Bolded values indicate the top-3 models. * = Flattening pooling

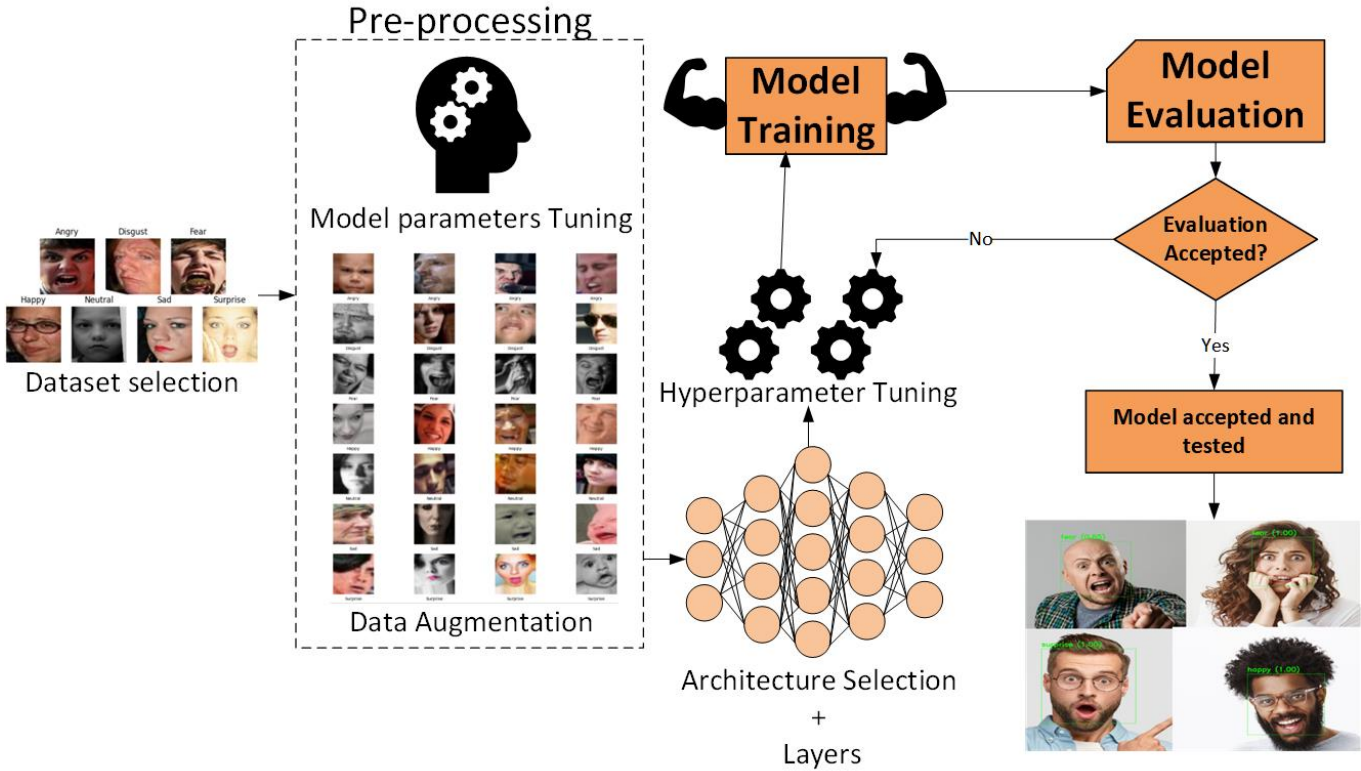

Fig. 2: Protocol Overview including pre-processing, model training and testing.

## III. PRACTICAL PROTOCOL

### A. Protocol overview

Besides the BTFER dataset, a comprehensive protocol for practical FER evaluation is proposed. Our model development and testing protocol is depicted in Fig. 2. Specifically, given a dataset, we first conduct data augmentation to enhance the model parameter tuning in the following steps. Then, the preprocessed dataset will be fed into every network architecture with hyperparameter adjustment. Next, a set of training settings will be assigned for advanced evaluation within the model training. Finally, the trained models will be tested on the BTFER dataset to evaluate their practical FER performance. The protocol was applied for all model architectures and datasets to enable uniform testing conditions.

### B. Public network architectures

We tested a total 23 CNN model architectures. These are listed on Tables 2 & 3. The majority of these architectures are available on the TensorFlow Keras library, others have been provided with related research papers (i.e., RepVGG, LeNet5, GoogLeNet). In the following, we provide a short description of all models. More details and Python implementations of the models can be found in the repository https://github.com/gipinze/ResearchModels_GIT.

**VGG16:** A deep CNN that has 16 layers, including 13 convolutional layers, including multiple convolutional and max-pooling layers, followed by fully connected layers [16].

**VGG19:** It is an extension of VGG16, maintaining a similar architecture but with additional convolutional layers. VGG19 offers improved representational capacity but with more expensive computationally [16].

**VGGNet:** It is known for its uniform architecture and deep structure. It popularized the use of deep convolutional neural networks in computer vision tasks and served as a benchmark for subsequent models [16].

**MobileNet:** A lightweight CNN designed for efficient inference on mobile and edge devices. It employs depth-wise separable convolutions to reduce computational complexity while maintaining good accuracy [17].

**MobileNetV2:** An improved version of the original MobileNet architecture. It introduces inverted residual blocks and linear bottlenecks, leading to improved efficiency and performance compared to the original MobileNet [17].

**EfficientNetV2_B0:** EfficientNetV2 is a family of CNN models developed to achieve excellent accuracy and efficiency. EfficientNetV2_B0 is the base model in this family, featuring compound scaling, which optimizes depth, width, and resolution to achieve a balance between accuracy and computational efficiency [18].

**NasNet Mobile:** A neural architecture search-based CNN that is optimized for mobile devices. It utilizes a cell-based structure with skip connections and achieves competitive accuracy with reduced computational complexity [19].

**ResNet50:** A variant of the ResNet (Residual Network) architecture that has 50 layers and utilizes residual connections to alleviate the vanishing gradient problem [20].

**ResNet18:** A variant of the ResNet architecture with 18 layers. It provides a shallower and more lightweight version compared to deeper variants like ResNet50, making it suitable for scenarios with limited computational resources [20]. In our experimentation, we decided to add extra layers to improve the initial accuracy of ResNet18, these extra layers were Dropout (0.2) + Dense layer (1024) with ReLU activation and one last Dropout layer before the SoftMax, to prevent overfitting and achieve higher accuracy.

**DenseNet121:** A densely connected CNN architecture where each layer is connected to every other layer in a feed-forward manner. This connectivity pattern promotes feature reuse, reduces the number of parameters, and improves gradient flow, leading to efficient and accurate models [21].

**LeNet5:** One of the early CNN architectures was primarily designed for handwritten digit recognition and played a pivotal role in popularizing CNNs. LeNet5 consists of convolutional, pooling, and fully connected layers [22].

**AlexNet:** It consists of five convolutional layers, max-pooling layers, and fully connected layers, showcasing the power of deep learning on large-scale image classification tasks [23].

**RepVGG:** RepVGG is a recent CNN architecture that introduces a new design principle called Re-parameterized CNN, which unifies the training and inference processes. It achieves competitive accuracy while providing simplicity and interpretability [24].

**GoogLeNet:** It is a type of CNN based on the Inception architecture. It utilizes Inception modules, which allow the network to choose between multiple convolutional filter sizes in each block [25].

**Xception:** The name stands for "Extreme Inception" and it was inspired by Inception Architecture. Instead of using standard convolutions, Xception uses depth-wise separable convolutions and aims to capture local and global dependencies [26].

**InceptionV3:** It is a CNN architecture from the Inception family. It makes several improvements including using Label Smoothing, factorized 7 x 7 convolutions, and the use of an auxiliary classifier to propagate label information lower down the network [25].

**SeNet:** Squeeze-and-Excitation Networks (SeNet) is a CNN architecture that employs squeeze-and-excitation blocks to enable the network to perform dynamic channel-wise feature recalibration [27].

**ZF_NET:** It comprises convolutional, pooling, and fully connected layers, demonstrating the effectiveness of deep networks for image classification tasks.

C. *Self-designed network architecture*

In addition to public networks, we also propose several self-designed architectures to enrich the benchmark protocol.

**Sequential 3 Conv:** Our Sequential 3 Conv model consists of:
- The first layer is a Conv2D layer with 32 filters, (3, 3) kernel size, ReLU activation, and 'same' padding.
- The second layer is another Conv2D layer with 64 filters and (3, 3) kernel size, followed by ReLU activation.
- MaxPooling2D reduces spatial dimensions with (2, 2) pooling.
- Dropout at a rate of 0.25 helps prevent overfitting.
- Two more Conv2D layers with 128 filters and (3, 3) kernels, each followed by ReLU, are added. MaxPooling2D and Dropout follow the second Conv2D.
- A Conv2D layer with 512 filters, (3, 3) kernel, ReLU activation, and Batch Normalization is applied. MaxPooling2D and Dropout follow.
- GlobalAveragePooling2D reduces spatial dimensions.
- A Dense layer with 1024 units and ReLU activation is added.
- Dropout at a rate of 0.5 is applied to the Dense layer.
- The final Dense layer with 7 units (for class classification) uses SoftMax activation for output probabilities.

**Sequential 4 Conv:** Our Sequential 4 Conv model consists of:
- The first layer is a Conv2D with 32 filters, (3, 3) kernel, ReLU activation, and 'same' padding.
- The second Conv2D layer has 64 filters and (3, 3) kernel with ReLU activation.
- MaxPooling2D reduces spatial dimensions using (2, 2) pooling.
- Dropout with a rate of 0.25 helps prevent overfitting.
- Two more Conv2D layers with 128 filters and (3, 3) kernels follow, each with ReLU activation. MaxPooling2D and Dropout are applied after the second Conv2D.
- Two additional Conv2D layers have 256 filters and (3, 3) kernels, each followed by ReLU activation. MaxPooling2D and Dropout are applied after each of these layers.
- Another Conv2D layer with 512 filters, (3, 3) kernel, ReLU activation, and Batch Normalization is added.
- Dropout at a rate of 0.2 is applied.
- Data is flattened using the Flatten layer.
- A Dense layer with 1024 units and ReLU activation is added.
- Dropout with a rate of 0.5 is applied to the Dense layer.
- The final Dense layer with 7 units uses SoftMax activation for classification probabilities.

**Sequential Simple:** Our Sequential Simple model consists of:
- The first layer is a Conv2D with 32 filters, (3, 3) kernel, ReLU activation, and input shape (48, 48, 3).
- A second Conv2D layer has 64 filters and (3, 3) kernel with ReLU activation.
- MaxPooling2D with (2, 2) pooling reduces spatial dimensions.
- Dropout at a rate of 0.1 helps regularize the model.
- The process repeats with a Conv2D layer (128 filters, (3, 3) kernel), followed by MaxPooling2D and Dropout.
- Another Conv2D layer (256 filters, (3, 3) kernel), followed by MaxPooling2D and Dropout.
- Output is flattened using the Flatten layer.
- A Dense layer with 512 units and ReLU activation is added.
- Dropout at a rate of 0.2 is applied for regularization.
- The final Dense layer with 7 units uses SoftMax activation for classification.

**Ensemble:** An ensemble CNN model combines the predictions of multiple individual CNN models to improve overall performance. It leverages diverse architectures or variations in model parameters to capture different aspects of the data and make more accurate predictions. Our Ensemble model comprises three CNN models (x1, x2, x3), each following a similar structure: Convolutional layers, Batch Normalization, Activation functions, Max Pooling, Dropout, Global Average Pooling, and Dense layers.

- For x1, it has three Convolutional layers with 32, 64, and 128 filters, a (3, 3) kernel size, followed by Batch Normalization and ReLU Activation. Max Pooling with (2, 2) pool size and 0.25 dropout is applied twice. After Global Average Pooling, two Dense layers with 512 units and 0.5 dropout are added.
- The output of x1 is passed through a Dense layer with SoftMax activation for classification.
- x2 and x3 follow similar patterns with variations in Convolutional layers.
- Finally, the outputs of all three models are averaged to produce the ensemble's output.

**Modular:** This CNN architecture is composed of modular building blocks, allowing flexibility in designing and incorporating different components or subnetworks for specific tasks or domains. It offers a modular and customizable approach to constructing CNN models. Our modular architecture consists of a local feature sub-network and a global feature sub-network. The local feature sub-network includes three convolutional layers, three max-pooling layers, two fully connected layers, and two dropout layers. These layers extract local features from the input images. The global feature sub-network is similar but uses larger kernel sizes in the convolutional layers. It also has three convolutional layers, three max-pooling layers, two fully connected layers, and two dropout layers. The outputs of the two sub-networks are then concatenated. The model has a final dense layer with SoftMax activation for multi-class classification. Overall, the model has a total of 24 layers, including convolutional, max pooling, fully connected, dropout, and concatenate layers.

**Single model multi-branch (SMMB):** A single model multi-branch CNN architecture consists of a single network with multiple branches, where each branch specializes in extracting features or making predictions for specific subtasks or classes. It enables the network to learn diverse representations and perform multiple tasks simultaneously. Our Single model multi-branch has two branches: face detection and emotion recognition.

- Shared convolutional layers include three Conv2D layers with 32, 64, and 64 filters, using (3, 3) kernels and ReLU activation.
- The face detection branch has MaxPooling2D, Conv2D (128 filters), Flatten, and Dense (64 units) layers.
- The emotion recognition branch has Conv2D (128 filters), MaxPooling2D, Conv2D (256 filters), MultiHeadAttention, LayerNormalization, Dense (128 units), and Dropout (0.5) layers, followed by Flatten.

Both branches are merged using the concatenate layer. The output layer has a Dense layer with 7 units and SoftMax activation for emotion recognition. The Model class is used for model creation, specifying input and output.

D. *Class Balancing managements.*

As mentioned before, public datasets used for FER usually have a notable imbalance between their classes. It can cause overfitting and hinder the generalization of the model during deployment. To solve the class imbalance issue, several options exist, including oversampling, under-sampling, and class weighting.

Class weighting is a technique in machine learning where each class in a classification problem is assigned a weight or importance factor, which is the most common strategy in existing FER studies. These weights are used during the training process to adjust the impact of each class on the model's loss function or optimization algorithm. Class weights are typically assigned based on the class distribution in the training dataset, with higher weights given to minority classes and lower weights to majority classes. The purpose of class weighting is to mitigate the effects of class imbalance and improve the model's ability to make accurate predictions for all classes in the presence of imbalanced data.

In this work, we applied two class weighting techniques, i.e., simple class weighting and conservative class weighting.

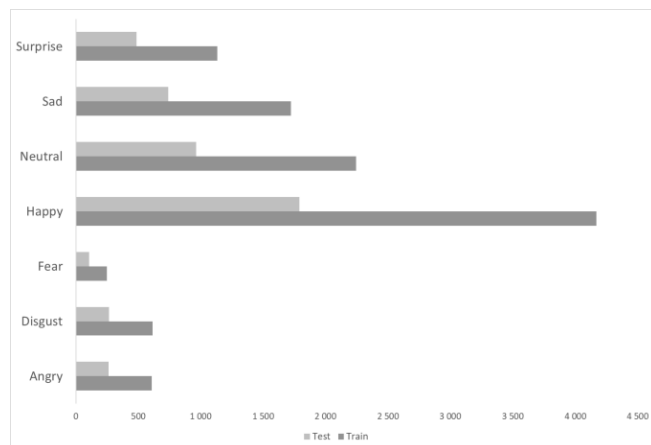

Fig 3: RAF-DB natural classes balance

**Simple Class Weighting:** This addresses class imbalance helping the model to be more sensitive regarding the less represented classes, in highly imbalance datasets this approach can lead to overfitting due to this reason, this approach is more effective with more balanced datasets, like AffectNet.

To perform our simple class weighting we implemented the following code:

```python
# Compute the class weights
class_to_idx=validation_generator.class_indices
idx_to_class={class_to_idx[cls]:cls for cls in class_to_idx}
print(idx_to_class)

(unique, counts) = np.unique(train_generator.classes, return_counts=True)
cw=1/counts
cw/=cw.min()
class_weights = {i:cwi for i,cwi in zip(unique,cw)}
print(class_weights)
```

Fig 4: Simple Class Weighting Python code

Specifically, we perform simple class weighting as the following steps:

1. First, we create a dictionary (class_to_idx) that maps class labels to their corresponding integer indices. This mapping helps us understand the numerical representation of class labels, which can be useful for various purposes, including printing the dictionary with idx_to_class.
2. Next, we calculate the number of images per class in the training dataset. This information provides us with the class distribution in the data, which is crucial for addressing class imbalance.
3. Finally, we assign the number 1 to the highest-represented class (the class with the most samples) and then compute class weights. These class weights are determined based on the relative imbalance between classes. We give more weight to underrepresented classes by dividing the counts of the most common class by the counts of each class. The result is that classes with fewer samples receive higher weights, while the most common class has a weight of 1.

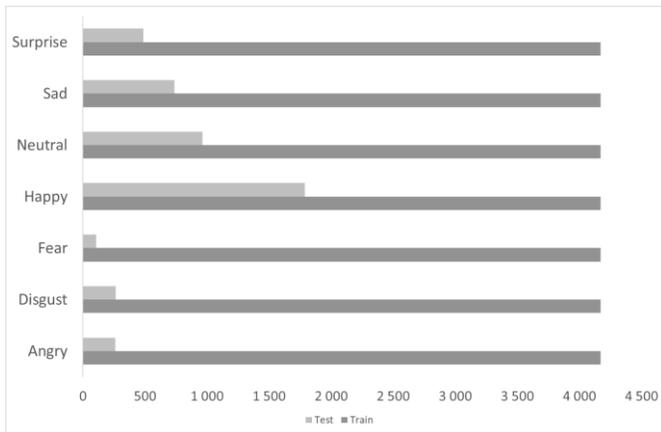

Fig 5: RAF-DB Simple class weighting balance

| Class name | Class index | weight |
|---|---|---|
| Angry | 0 | 6.88 |
| Disgust | 1 | 6.80 |
| Fear | 2 | 16.81 |
| Happy | 3 | 1.00 |
| Neutral | 4 | 1.86 |
| Sad | 5 | 2.42 |
| Surprise | 6 | 3.68 |

Fig 6: RAF-DB simple class weighting dictionary

**Conservative Class Weighting:** This class weighting technique aims to address class imbalance by encouraging the model to pay more attention to minority classes during training without allowing the weights to become excessively large, which could lead to overfitting. It strikes a balance between mitigating class imbalance and preventing potential overfitting. To perform our conservative class weighting we implemented the following code:

```
from collections import Counter
counter = Counter(train_generator.classes)
max_val = float(max(counter.values()))
class_weights = {class_id : np.minimum(max_val/num_images,3) for class_id, num_images in counter.items()}
```

Fig 7: Conservative class weighting Python code

This code calculates class weights based on the class distribution in the training data. It first counts the number of samples per class using the Counter class. Then, it computes class weights by comparing the number of samples in each class to the most common class. The weights are capped at a maximum value of 3 to prevent extreme values. These class weights are useful for training machine learning models, especially when dealing with class imbalance. We found this approach more useful for highly imbalanced datasets like RAF-DB and less significant for AffectNet.

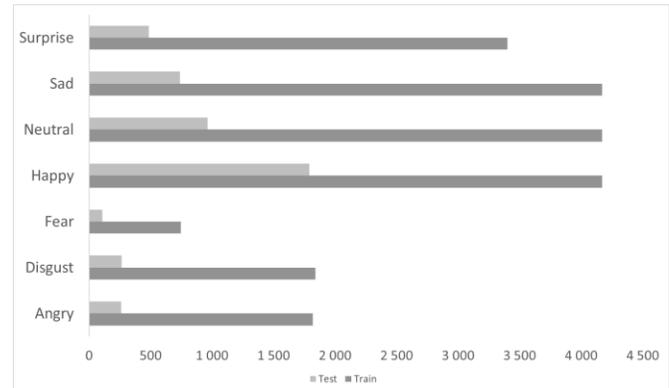

Fig 8: RAF-DB Conservative Simple class weighting balance

| Class name | Class index | weight |
|---|---|---|
| Angry | 0 | 3.00 |
| Disgust | 1 | 3.00 |
| Fear | 2 | 3.00 |
| Happy | 3 | 1.00 |
| Neutral | 4 | 1.86 |
| Sad | 5 | 2.42 |
| Surprise | 6 | 3.00 |

Fig 9: RAF-DB conservative class weighting dictionary

E. *Pretrained*

Models pretrained on large-scale datasets can significantly improve the performance in downstream tasks by effective finetuning. Commonly used pre-trained dataset in FER studies are ImageNet and VGGFace.

**ImageNet** is an image database designed to train different convolutional neural networks, in our case, the models use ImageNet-1k as pre-trained weight, which is equivalent to ImageNet with 1000 classes, these classes are varied and include elements such as animals, objects, means of transport among others.

**VGGFace** is a dataset used for face identification tasks. It is possible to use VGGFace pre-trained weights by adding the Keras-VGGFace library to our notebook and calling VGG16 as model for training. With this configuration we can use VGGFace as pre-trained weights.

Apart from the finetuning, training from scratch makes the model learn from zero, without using any pre-trained weight.

F. *Training configurations*

Different from current FER method using data augmentation during the training phrase, in this work, we followed Keras guidelines of pre-processing images. These recommendations included re-scaling (1/255), rotation range (20), width shift

(0.2), height shift (0.2), zoom range (0.2) and horizontal flip (True) [28].

To ensure a fair comparison, same hyper parameters are applied for all models: Learning rate (1e-4), the same reduction of learning rate based on validation loss, with a patience of 10 epochs, with a factor of 0.50, minimum learn rate of 1e-10 and early stopping with a patience 20 epochs.

For the training process, we also follow the recommendations of TensorFlow Keras, regarding the pre-processing, freezing and fine-tuning models. Using pre-trained weights and freezing the last interconnected layers, adding the same number of new layers, and training the model for 30 epochs or until the model does not learn anymore (early stopping strategy).

After the first round of training, we selected the best model (based on our checkpoint), unfreeze the whole model including the recently added layers and proceed to train the whole model until the training stops by early stopping. This is important because it allowed us not to fall into overfitting and obtaining the best out of the model in case of not being able to learn more during training..

Methods for the facial recognition of emotions are varied. There are ready-to-use tools such as DeepFace and FaceApi that allow us to analyze facial expressions in a simple way and without the need to train the model already trained by other developers.

## IV. EXPERIMENTS

Our objective in this paper was to study, analyze and test open-source models for FER performance, mainly focusing on models available in TensorFlow Keras, which are among the most popular in the developer community. In this section, we conducted experiments to explore different factors influencing the accuracy of Facial Expression Recognition CNN models. We developed a uniform testing protocol that allowed us to evaluate 17 network architectures for their performance.

### A. Metrics

We evaluated the models using different metrics, which included the following.

**Accuracy:** To measure the correct classification of the model, this metric calculates the ratio between correctly classified samples against the total number of samples (Tables 2 & 3).

**Confusion Matrix:** Confusion Matrix provides clearer information regarding the model's classification performance for each class. It shows the combination of true positives, true negatives, false positives, and false negatives.

**Test Accuracy (Out-of-sample):** For practical development applications of FER, it's not enough to test models using the same dataset used during training, this might show us that our model performance is high, while might have problems to generalize (problems to work in the real world). Based on the even distribution of our collected BTFER dataset, it is possible to test performance of both public and self-designed models encountering unseen data.

### B. Implementation

**Platform:** Our implementation and testing platform was local based with the following specs:
- Windows 11
- 2 GPU Nvidia 3060 (12Gb x 2 gpus)
- 64gb RAM at 3200 MHz
- Python 3.9
- TensorFlow 2.10

**Hyperparameters:** As mentioned above, the hyperparameters for our model training were the same for all the models.
- Data Augmentation parameters
- Learning Rate: 0.0001 (1e-4)
- Batch size: 32 x Strategy (2 GPU)
- Beta_1: exponential decay of 0.9
- Beta_2: exponential decay of 0.999
- Patience: 10 epochs
- Factor: 0.50
- Minimum Learning Rate: 1e-10

**Network Layers:** For all the pre-trained models, we have applied the same strategy. We froze the fully connected layers and added new layers with the same parameters:
- Base Model: The selected architecture with its fully connected layers frozen
- Pooling Techniques: We use 2 different pooling techniques, GlobalAveragePooling2D and Flatten. In Tables 2 & 3, we indicate models and which of these techniques were used.
  a) GlobalAveragePooling2D computes the average of the existing values across the spatial dimensions not increasing the number of parameters.
  b) Flatten converts multi-dimensional tensors into a single one-dimensional vector by stacking (compressing or flattening) its contents.
- Dense layer with 256 neurons and ReLU
- Dropout (0.2)
- Dense layer with 128 and ReLU
- Dense layer with num_classes (7) and activation SoftMax for classification.

### C. Performance Comparison

**Different input resolutions:** We used two different approaches regarding the resolution of the images. This was to corroborate the fact that when working with the recognition and classification of images, a higher resolution should mean greater precision. We wanted to compare how similar models behave against the same datasets but at different resolutions.

Tables 1 and 2 showed a wide variety of results, where we can see how significant the correct use of a well-labelled dataset is (with RAF-DB being the dataset with the best accuracy). It could be replicated when comparing the models, through cross-validation, with our own dataset (BT-FER).

**Different pre-trained weights:** We were interested in comparing pre-trained weights against those trained from scratch. For this, we evaluated the accuracy of the models using all the datasets available in our study, using three different initial states and training VGG16 model. We used pre-trained weights of VGGFace in Keras from the Keras-VGGFace library, allowing us to train on the VGG16 architecture with three different starting points and with two different resolutions, i.e., 48 x 48 and 224 x 224, respectively.

D. *Experiment 1 – Different pre-trained weights, different performance?*

In this experiment, we aim to see how much difference the usage of pre-trained weights affects the models' accuracy. We trained the VGG16 model with three different starting weights, i.e., ImageNet (IN), VGGFace (VF), and from Scratch (SC).

**48 x 48 Resolution**: When training models with low resolution images, i.e., 48 x 48, the best approximation was using ImageNet as pre-trained weights, secondly was training the model from scratch with random initial weights, and thirdly using VGGFace pre-trained weights [29] in the corresponding library, as shown in Fig. 10 and 11.

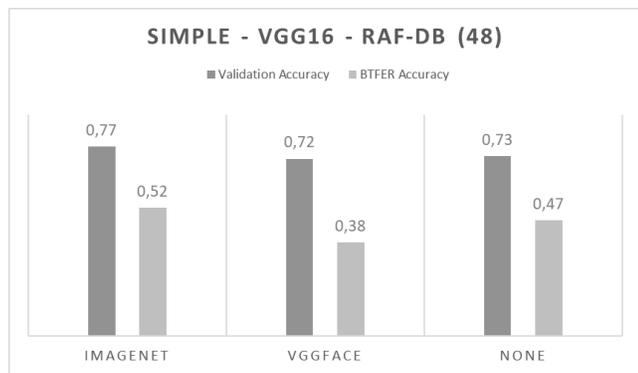
Fig. 10: Pre-train weights accuracy difference – RAF-DB Simple 48 x 48

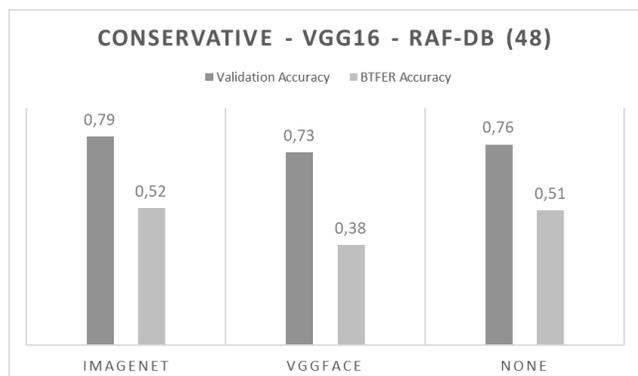
Fig. 11: Pre-train weights accuracy difference – RAF-DB – Conservative 48 x 48

This difference in performance can be attributed to ImageNet's ability to generalize across various elements. In contrast, VGGFace has been specifically trained on facial data, which often necessitates higher resolution for effective feature identification. Consequently, VGGFace demonstrates reduced generalization capabilities. The effect of the sampling strategy was negligible with conservative sampling scoring at best 0.03 over simple class weighting.

**224 x 224 resolution:** When we trained the same models with higher resolution (224 x 224), we were able to visualize a greater influence of the pre-trained weights (VGGFace), not only in the validation precision, but also against our dataset (BTFER). Results are depicted in Fig. 12 and 13.

This difference is because the weights pretrained with VGGFace are considerably more like our dataset, being both faces. By working with higher resolution, it is possible to extract a greater and better number of features. of the images.

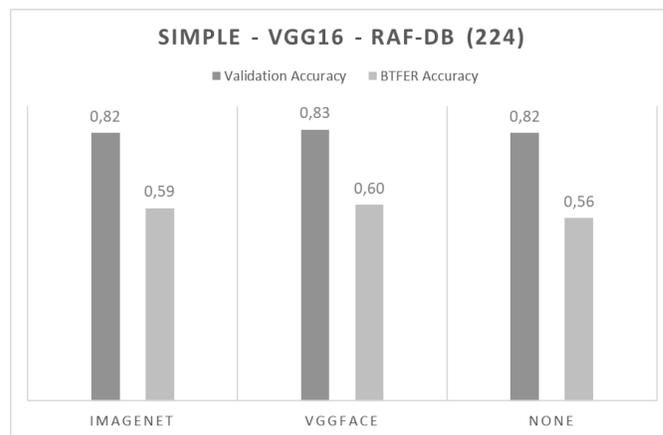
Fig. 12: Pre-train weights accuracy difference – RAF-DB Simple 224 x 224

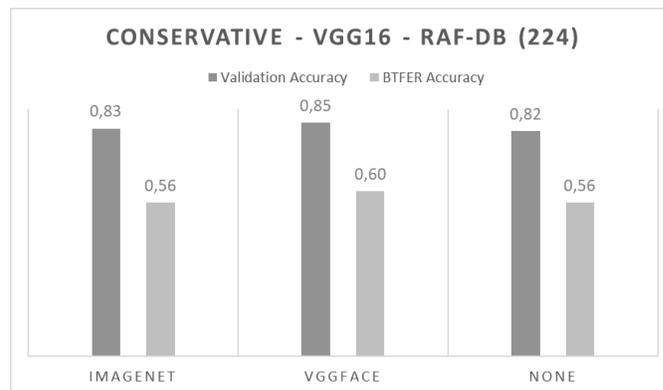
Fig. 13: Pre-train weights accuracy difference – RAF-DB – Conservative 224 x 224

E. *Experiment 2 – Less parameters, good performance?*

In practical settings, sufficient computation resources are sometimes unavailable. Thus, it is necessary to further explore how to obtain high accuracy with smaller models.

**48 x 48 Resolution:** According to the experimental results, it was possible to obtain high accuracy (Over 70%) using CNN models with less parameters, as seen in Fig. 14. The best results was obtained using RAF-DB with both resolutions.

In table 4, we can see the model with the best performance was the ResNet18. Additionally, those ones using VGG as backbone obtained the best performance. One possible explanation is that the VGG architecture can extract features from images even with noise or defective quality [5].

Alternatively, if we are willing to sacrifice accuracy for speed, then we can use smaller models like Mediapipe + VGGNet or a Sequential model with four convolutional layers (6[th] place), which might give us good accuracy as well.

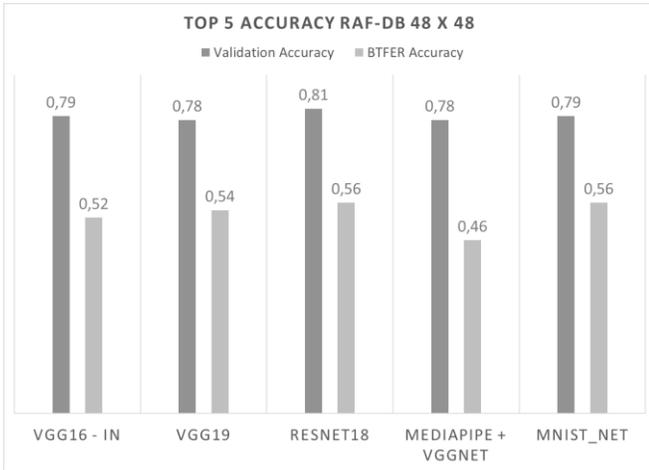

Fig. 14: Accuracy comparison for top-5 models with less parameter on RAF-DB – 48 x 48

| Model | Num Parameter (In Millions) | Validation Accuracy | BTFER Accuracy |
|---|---|---|---|
| VGG16 - IN | ~15M | 0,79 | 0,52 |
| VGG19 | ~20M | 0,78 | 0,54 |
| ResNet18 | ~12M | 0,81 | 0,56 |
| MediaPipe + VGGNet | ~5,5M | 0,78 | 0,46 |
| MNIST_NET | ~18M | 0,79 | 0,56 |

Table 4: Accuracy comparison table, top 5 best models RAF-DB – 48 x 48 (Simple Class weighting)

**224 x 224 Resolution:** When we tested the models with bigger resolution (224 x 224), we found that the models with outstanding performance were those with trained using faces (VGGFace). As shown in Fig. 15 and Table 5, the five most outstanding models were all pre-trained with VGGFace.

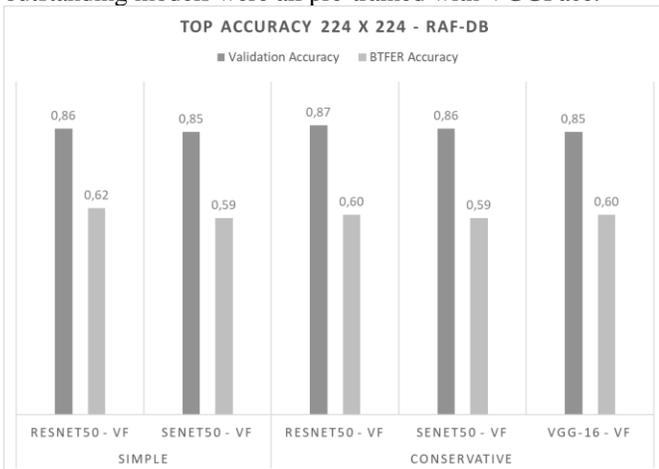

Fig. 15: Accuracy comparison for top-5 models with less parameter on RAF-DB – 224 x 224

| Balance | Model | Validation Accuracy | BTFER Accuracy |
|---|---|---|---|
| Simple | ResNet50 - VF | **0,86** | **0,62** |
|  | SeNet50 - VF | **0,85** | **0,59** |
| Conservative | ResNet50 - VF | **0,87** | **0,60** |
|  | SeNet50 - VF | **0,86** | **0,59** |
|  | VGG-16 - VF | **0,85** | **0,60** |

Table 5: Accuracy comparison table, top 5 best models RAF-DB – 224 x 224 (Simple Class weighting)

F. *Experiment 3 – Importance of well-labeled datasets*

After training different models using three different datasets, we found that RAF-DB, even being a smaller dataset (~15K images) achieves better accuracy on average compared to FER2013 (~35K images) and AffectNet (~27K), as presented in Fig. 16-19. These results were observed using both class balance strategies. For both scenarios, RAF-DB gave better results. This result highlights the importance of accurate labeling [30].

When training the models with 224 x 224 images, we were able to see that, like with smaller resolution, RAF-DB obtained higher accuracy, but in this scenario, was AffectNet that obtained the second place, next being FER2013, this might be cause to image resolution, FER2013 it is in 48 x 48 as default, while AffectNet512 x 512, and this difference might compensate the labeling problems.

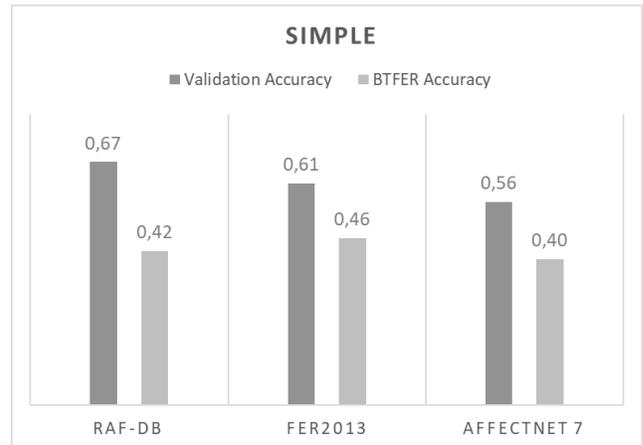

Fig. 16: Average accuracy datasets with simple class weighting balance 48 x 48

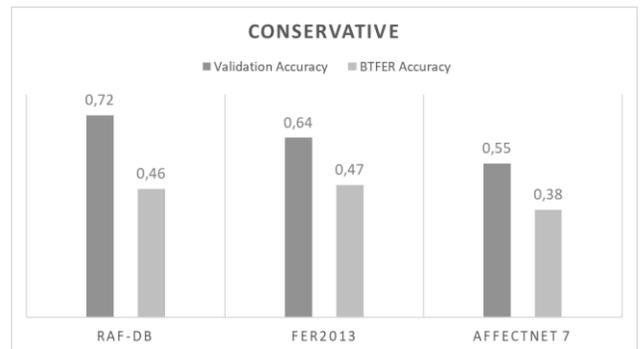

Fig. 17: Average accuracy datasets with conservative class balance 48 x 48

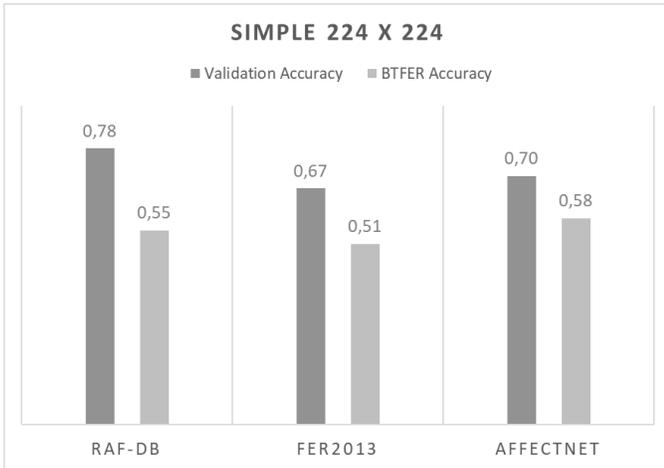

Fig. 18: Average accuracy datasets with simple class weighting balance 224 x 224

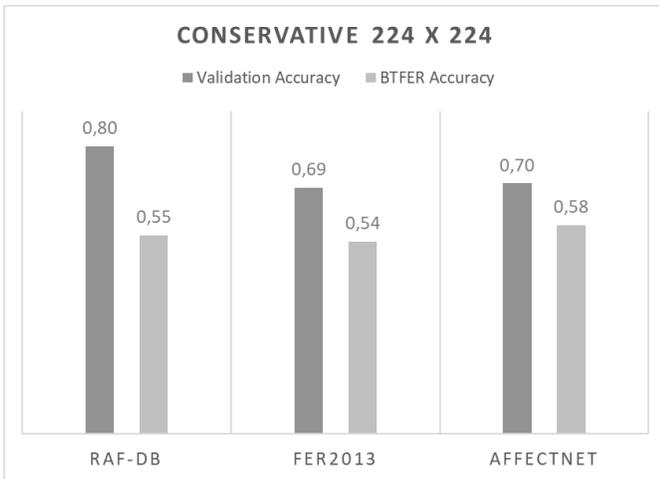

Fig. 19: Average accuracy datasets with conservative class balance 224 x 224

In summary, faces that we do not consider representations of certain emotions and that had been bad categorized will affect the model's generalization, an example of this might be a face labeled as happy, while we can agree that it might represent better a neutral expression.

G. *Experiment 4 – Class balance strategy affects more on models trained from scratch than those trained using pre-trained weights.*

**48 x 48 Resolution:** When we trained the models from scratch, the influence of our class balance strategy, in average, is more significant (as shown in Fig. 20) with those models trained from scratch than with those models trained using pre-trained weights (as shown in Fig. 21). This result is due to the lack of previous learning. A model training from scratch must learn all the features and representations, adding to this a class balance strategy might affect the mode due to overfitting. In other words, if a model is learning from scratch, a strategy that works considering overfitting might obtain better results, no matter what kinds of rebalancing strategies it used (Conservative vs. Simple weighting balance).

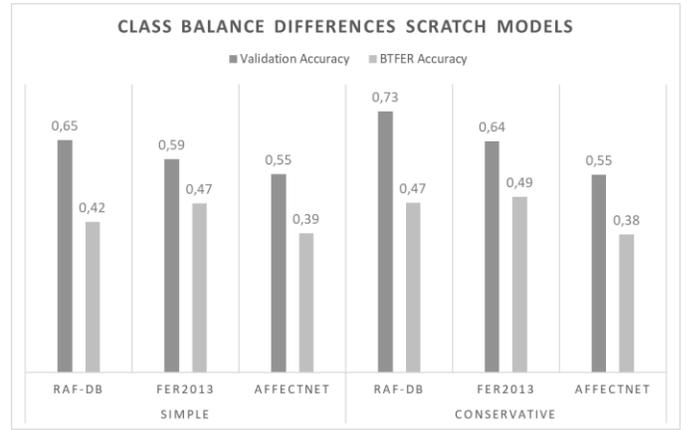

Fig 20: Differences accuracy models (avg) with no pre-trained weights (48 x 48)

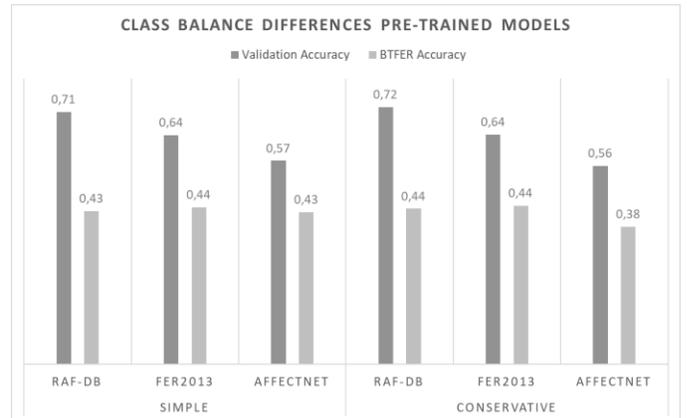

Fig 21: Differences accuracy models (avg) with pre-trained weights (48 x 48)

**224 x 224 Resolution:** On the other hand, we have models trained with a higher resolution in similar setups, as shown in Fig. 22 and 23. The difference between the models with simple weighting class balance has been reduced, if we compared it with 48 x 48 resolution, we could agree that, when training models with higher resolution the class balance strategy is less significant than a well labeled and balance dataset.

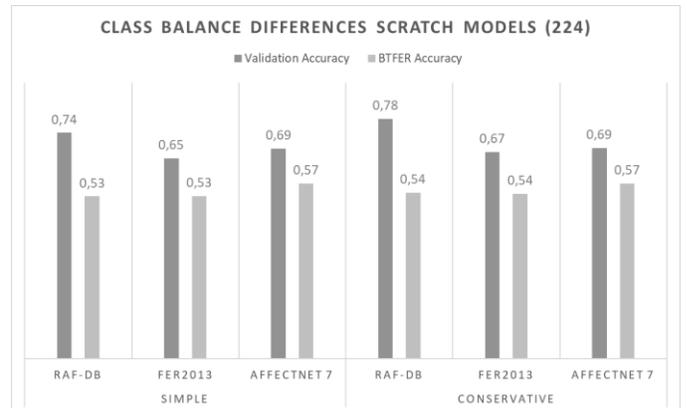

Fig 22: Differences accuracy models (avg) with no pre-trained weights (224 x 224)

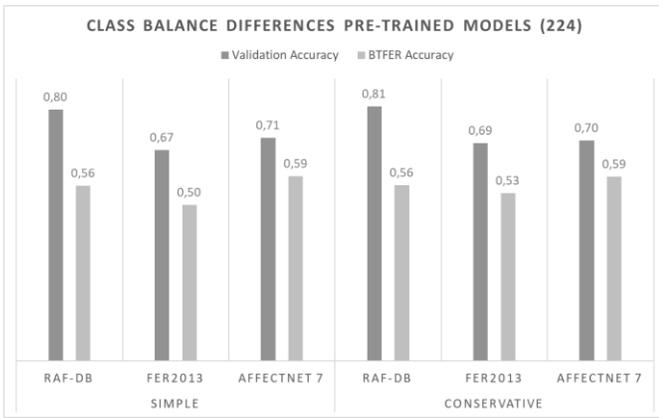

Fig 23: Differences accuracy models (avg) with pre-trained weights (224 x 224)

H. *Experiment 5 – Higher resolution equals more accuracy*

When we compare the same models trained with 48 x 48 resolution images against those trained with 224 x 224 images, we found that the accuracy difference between them were ~0.06 and ~0.09 for validation and testing, respectively, as depicted in Fig. 24. Hence, by increasing the resolution, accuracies were increased by ~9% and ~19%, respectively. This was expected as there are more facial details available with higher resolution, which demonstrated significant improvement in FER accuracy when the image resolution was higher, having 60.45% with 48 x 48 while 74.57% with 400 x 400 [31].

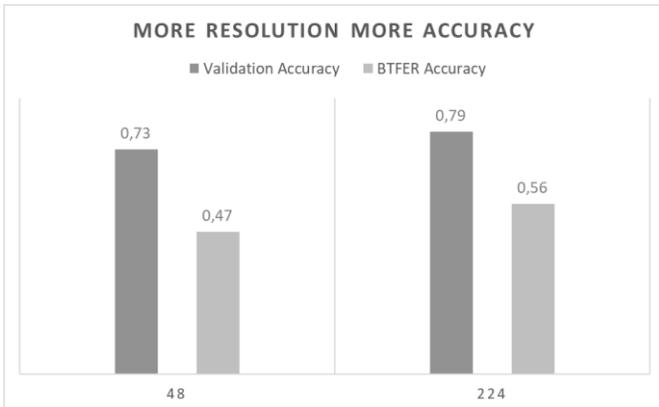

Fig 24: Accuracy differences between models trained with 224 and 48 resolution images (RAF-DB)

I. *Experiment 6 – Best model for FER*

Based on the results, we divided our best models into 2 groups, based on images resolutions. As we can see in Table 6, when working with low resolution images (48 x 48), the best option to obtain high accuracy is to work with ResNet18 with two dropout and one heavy dense layer, another option, and easier for new users, might be working with VGG16 + ImageNet (as shown in Table 1) as pre-trained weights might give an acceptable result without building a ResNet18 model from scratch.

Alternatively, if we want a fast implementation to be used in edge-devices (e.g., Raspberry Pi), we can develop sequential models, like our Sequential model with 4 convolutional layers and flatten as pooling layer to obtain small architectures with descent accuracy.

In order to use efficiently models trained with TensorFlow, we should convert them using TensorFlow Lite [32], with this conversion we reduce the model's size complexity making it more suitable for deployment in Edge computing devices such as raspberry pi, additionally TFLite provides hardware acceleration for inference on devices that supports it, improving the model's performance [32].

For the 224 x 224 resolution, the highest accuracy was obtained using ResNet50 with VGGFace as pre-trained weights [29]. While if we aim for a faster model, able to be implemented in edge computing devices, our best option is the well-known DenseNet121, which get ~5% less accuracy, but it is still very reliable.

|  | Model | Num Parameter | Validation Accuracy | Test Accuracy |
|---|---|---|---|---|
| 48 | ResNet18 + Add | ~ 12M | 0,81 | 0,55 |
|  | Sequential 4 Conv | ~3 M | 0,76 | 0,54 |
| 224 | ResNet50 - VF | ~24M | 0,85 | 0,62 |
|  | DenseNet121 | ~7M | 0,82 | 0,60 |

Table 6: Best performance models with 48 x 48 and 224 x 224

J. *Discussion*

This research aims to create impact to the field of FER by unraveling crucial insights into the intricate dynamics that influence model accuracy and performance.

Through a series of meticulously experiments, the study has not only shed light on the optimal strategies for FER model design and training across varying image resolutions but has also unearthed nuanced relationships between pre-trained weights, dataset quality, and class balance strategies.

The research's significance lies in its ability to guide both researchers and practitioners in making informed decisions, facilitating the creation of more accurate and efficient FER models tailored to specific contexts. By offering a comprehensive roadmap for model selection, dataset curation, and resolution considerations, this work serves as a valuable cornerstone for advancing the state-of-the-art in FER, fostering advancements that can be leveraged across fields ranging from human-computer interaction to emotional AI applications.

V. PRACTICAL APPLICATIONS

A. *TF-Lite and edge devices*

The practical applications from this research are abundant, particularly in scenarios where real-time and efficient Facial Expression Recognition (FER) is essential. Leveraging the insights gained from the experiments, deploying FER models on edge devices like the Raspberry Pi using TensorFlow Lite becomes a feasible and impactful avenue. With the identification of optimal models for different resolutions, such as ResNet18 for 48 x 48 images and ResNet50 with VGGFace pre-trained weights for 224 x 224 images, these models can be fine-tuned and converted to TensorFlow Lite format. This allows for their deployment on resource-constrained platforms like the Raspberry Pi, ensuring that even devices with limited computational power can accurately recognize facial expressions in real-time.

| Papers | Researcher | Year | Datasets | Structure | Accuracy Highest (~) |
| --- | --- | --- | --- | --- | --- |
| **Patt-Lite** | Jia Le Ngwe, Kian Ming Lim, Chin Poo Lee, Thian Song Ong | 2023 | CK+, RAF-DB, FER2013, FER+ | CNN + Extraction Block | 95.05% |
| **ArBex** | Azmine Toushik Wasi, Karlo Šerbetar, Raima Islam, Taki Hasan Rafi, Dong-Kyu Chae | 2023 | Aff-Wild2, RAF-DB, JAFFE, FERG-DB, FER+ | ViT + CNN + MAN + AFN + RBM | 92.47% |
| **Poster++** | Jiawei Maoy Rui Xuy Xuesong Yin* Yuanqi Chang Binling Nie Aibin Huang | 2023 | AffectNet (7 & 8), RAF-DB, CAER-S | CNN + ViT + Facial Landmark | 92.21% |
| **APVit** | Fanglei Xue, Qiangchang Wang, Zichang Tan, Zhongsong Ma, Guodong Guo | 2022 | RAF-DB, FER+, AffectNet, SFEW2.0, ExpW, Aff-Wild | CNN + AM + ATP Blocks + MLP He | 91.98% |

Table 7: Works with the highest accuracy in Facial Expression Recognition (as of 2023)

The lightweight option presented by DenseNet121 offers a practical solution for edge computing devices, striking a balance between accuracy and computational efficiency. This becomes particularly relevant for applications that require FER capabilities within IoT devices or embedded systems.

The research's emphasis on the influence of class balance strategies on scratch-trained models is crucial for edge scenarios, where data imbalance might be prevalent. By understanding that pre-trained models are less sensitive to class imbalance due to their prior learning, developers can focus on optimizing class balance techniques, enhancing model performance even in constrained environments. In essence, the findings of this research guide the development of efficient FER models that can seamlessly integrate into real-world applications, enabling a diverse range of use cases such as emotion-aware IoT devices, human-computer interaction systems, and sentiment analysis in educational settings.

### B. Ethical considerations in adopting FER technologies

This study reviewed FER only from a technological viewpoint but in real use cases ethical and legal issues need to be carefully considered as they vary between countries and applications.

The European Commission has taken facial recognition technology (FRT) under special review because it can be misused. Humans in private and public areas can be identified with FRT, which may violate the European Commission's data protection regulation (GDPR). The European Data Protection Board [33] released guidelines on the use of facial recognition technology in law enforcement in May 2023. The goal of the recently released guidelines proposition is to get public feedback, and the final guidelines will be updated and released later the basis of received expert and political feedback. The goal of the guidelines is to advise how FRT could be applied in order that it does not violate EU laws and regulations in Europe. Therefore, when researching and adopting FER, this future EU regulation on FRT should always be considered. It influences the requirement specification of FRT design and software features of FRT applications being developed. Therefore, more research is needed on what kind of features can be used to design and implement FRT applications that will not violate e.g., EU regulations and laws on privacy. The European Commission has taken a proactive role in regulating FRT and evidently many other countries or alliances will follow it.

### C. The future of the field

**Novel deep architectures:** The future of the FER field is poised for rapid advancements that build upon the foundations established by CNNs. While CNNs have demonstrated significant success in FER, the trajectory of the field is likely to encompass more sophisticated deep learning architectures. Approaches like Multi-Task CNNs and Vision Transformers, Multimodal implementations and attention mechanisms are anticipated to gain prominence, this is possible to visualize when looking at the most recent papers such as Patt-Lite [34], ArBex [11], Poster++ [9], APVit [8]. These advanced architectures can capture temporal dependencies and contextual information, enabling better feature extraction and understanding of facial expressions. However, the challenge lies in the increased computational demands of these complex models compared to traditional CNNs.

**Advanced training techniques:** Another promising avenue for future exploration is adaptative resolution techniques within CNN architectures for FER. As our research centered on fixed resolutions, the development of models that can adapt to varying resolutions could enhance accuracy by ensuring robust data utilization and generalization across different environments. Similar studies in other domains have sought to identify the ideal resolution for specific problems, and a similar approach could be adopted to optimize FER accuracy while managing computational resources efficiently [35]. Additionally, the pursuit of finding the optimal balance between computational power and accuracy in different image resolutions holds promise.

**Diverse datasets:** The future FER landscape should also prioritize the development of larger, balanced, and unbiased datasets to ensure models' robustness and generalization across diverse scenarios. Exploring other architectural paradigms like transformers and multi-task learning for FER could further enhance the field's capabilities. Besides, the notion of capturing micro-expressions to create a more nuanced representation of facial expressions is also gaining traction. Developing datasets that encompass a wide range of micro-expressions presents a challenge, yet addressing this could significantly boost the accuracy and applicability of FER models.

**Ethical considerations:** It is important to highlight that FER's applications extend beyond technological considerations. The real-world implementation of FER technologies necessitates careful attention to ethical and legal considerations, especially in the context of privacy, consent, and cultural variations. While our study has provided a technological perspective, the integration of FER into practical use cases should be approached with sensitivity to these broader implications. With advancements in technology, the FER field is set to make transformative strides, enhancing industries ranging from sports and retail to healthcare and security.

**Acknowledgment**

This research was supported by the AI Forum project funded by the Ministry of Education and Culture (OKM/116/523/2020). Dr. Yang Liu appreciated the support of the Finnish Cultural Foundation for North Ostrobothnia Regional Fund 'Towards Crowdsensing Facial Affect Encoder for Trustworthy Mental Wellbeing' (grant 60231712).